\documentclass[letterpaper, 10 pt, conference, final]{ieeeconf}   

\IEEEoverridecommandlockouts                              

\usepackage{graphics} 
\usepackage{amsmath, bm} 
\usepackage{amssymb}  
\usepackage{graphicx}
\usepackage{hyperref}
\usepackage{subcaption}
\usepackage{graphicx}
\usepackage{mathtools}
\usepackage{dsfont}
\usepackage{float}
\usepackage{makecell}
\usepackage{authblk}
\usepackage{algorithm}
\usepackage{algcompatible}
\usepackage{graphicx}
\usepackage{multirow} 
\usepackage{amsmath,amsfonts,bm}
\usepackage{bbm}
\usepackage{wrapfig}
\usepackage[table]{xcolor}
\usepackage[font=footnotesize,labelfont=bf]{caption}

\renewcommand{\baselinestretch}{1.0}

\newcommand{\numconnectors}{50}

\makeatletter
\renewcommand\AB@affilsepx{\hspace{1in} \protect\Affilfont}
\makeatother

\newcommand\blfootnote[1]{%
  \begingroup
  \renewcommand\thefootnote{}\footnote{#1}%
  \addtocounter{footnote}{-1}%
  \endgroup
}
\usepackage[noadjust]{cite}
\usepackage{subcaption}
\DeclareMathOperator{\E}{\mathbb{E}}

\def\methodname {domain adversarial information bottleneck{}}
\def\methodabbrv {DAIB{}}

\title{\LARGE \bf Learning on the Job: Self-Rewarding Offline-to-Online Finetuning for Industrial Insertion of Novel Connectors from Vision}

\author{Ashvin Nair$^{*1}$, Brian Zhu$^{*12}$, Gokul Narayanan$^{2}$, Eugen Solowjow$^{2}$, Sergey Levine$^1$}

\begin{document}


\vspace{-10pt}
\maketitle

\begin{abstract}
Learning-based methods in robotics hold the promise of generalization, but what can be done if a learned policy does not generalize to a new situation? In principle, if an agent can at least evaluate its own success (i.e., with a reward classifier that generalizes well even when the policy does not), it could actively practice the task and finetune the policy in this situation. We study this problem in the setting of industrial insertion tasks, such as inserting connectors in sockets and setting screws. Existing algorithms rely on precise localization of the connector or socket and carefully managed physical setups, such as assembly lines, to succeed at the task. But in unstructured environments such as homes or even some industrial settings, robots cannot rely on precise localization and may be tasked with previously unseen connectors. Offline reinforcement learning on a variety of connector insertion tasks is a potential solution, but what if the robot is tasked with inserting previously unseen connector?
In such a scenario, we will still need methods that can robustly solve such tasks with online practice. One of the main observations we make in this work is that, with a suitable representation learning and domain generalization approach, it can be significantly easier for the reward function to generalize to a new but structurally similar task (e.g., inserting a new type of connector) than for the policy. This means that a learned reward function can be used to facilitate the finetuning of the robot's policy in situations where the policy fails to generalize in zero shot, but the reward function generalizes successfully. We show that such an approach can be instantiated in the real world, pretrained on 50 different connectors, and successfully finetuned to new connectors via the learned reward function.
Videos and visualizations can be viewed at \href{https://sites.google.com/view/learningonthejob}{sites.google.com/view/learningonthejob}
\blfootnote{$^*$ First two authors contributed equally, $^1$ University of California, Berkeley. $^2$ Siemens Corporation. Correspondence:   \tt anair17@berkeley.edu}
\end{abstract}



\section{Introduction}

Generalizable policies require broad and diverse datasets, but for realistic applications, learning policies that can always generalize in zero shot to new objects and environments is often infeasible -- indeed, even humans do not exhibit such universal generalization capabilities. Instead, when faced with a task that we don't precisely know how to do, we can quickly learn the task by leveraging our prior knowledge and a little bit of practice. Reinforcement learning (RL) provides us with a way to implement this kind of \textit{learning on the job}, using online finetuning in the new domain or task, and potentially even extending it into a lifelong learning system where the robot improves its generalization capacity continually with each new task it masters.

However, instantiating this concept in a practical robotics setting requires overcoming a number of obstacles. The robot must be able to combine large amounts of diverse offline data with small amounts of targeted online experience, and do so in a way that doesn't require revisiting previously learned tasks or domains, which means that we need an offline RL algorithm that supports online finetuning. Perhaps more importantly, the entire finetuning process must be supported by the robot's own sensors, without privileged information or environment instrumentation, so as to retain the benefits of autonomous learning. In particular, this means that when adapting to a new task, the robot must be able to evaluate on its own whether it is making progress on the task, using a learned reward function.

\begin{figure}[b!]
    \vspace{-0.5cm}
    \centering
    \begin{subfigure}[b]{0.99\linewidth}
        \center
        \includegraphics[width=0.99\textwidth]{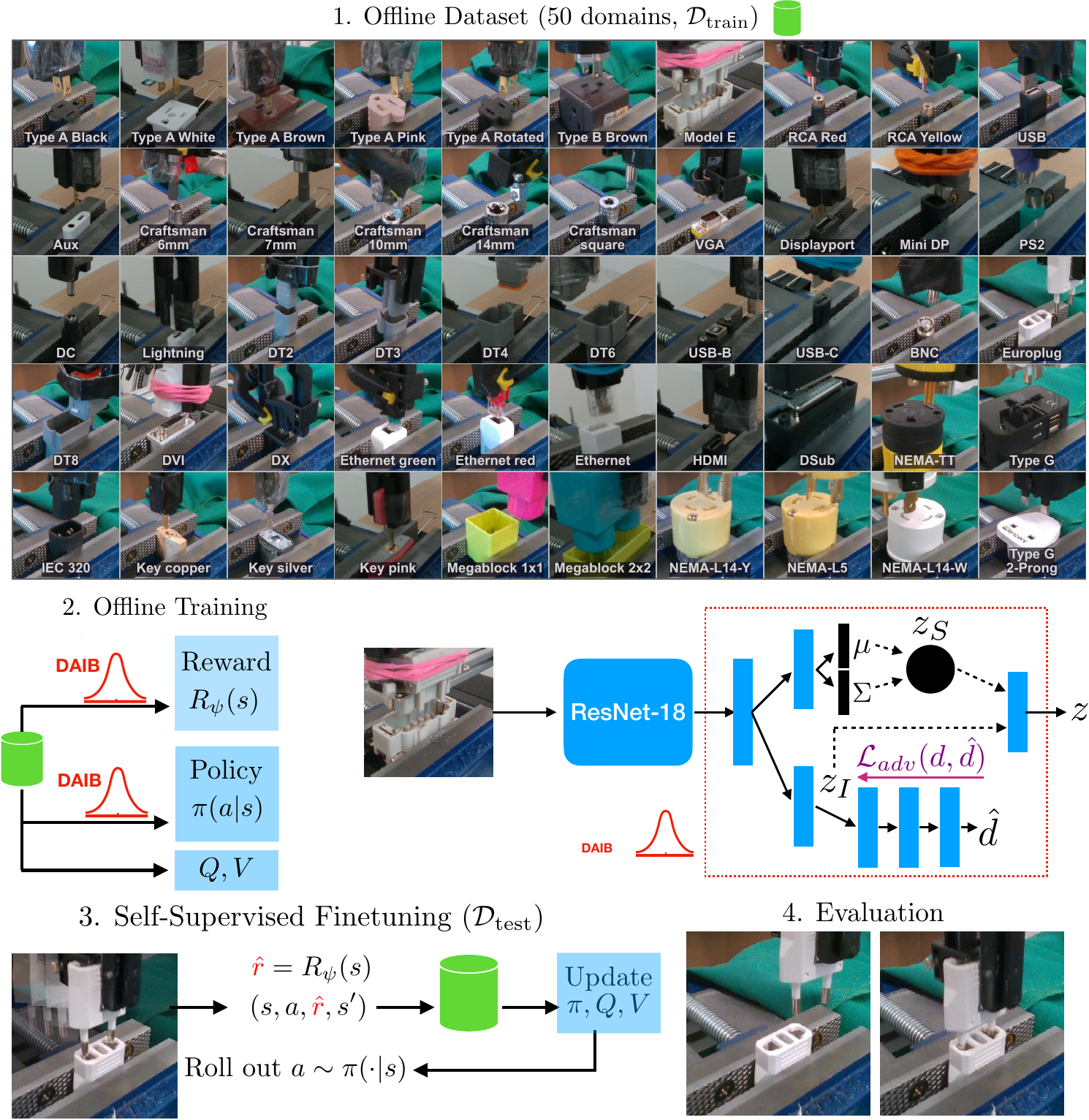}
    \end{subfigure}

    \caption{\footnotesize We describe how a robot can learn to insert an unseen connector from prior experience under realistic conditions by evaluating its own reward.
    (1) We first collect a diverse dataset with \numconnectors{} connectors. There is significant variation in the shape of the connectors, sockets, and background. (2) We learn a reward function, policy, and value functions using offline RL. We propose a domain adversarial information bottleneck (DAIB) in order to generalize to new domains. A domain invariance loss is applied on part of the latent representation $z_I$, while a domain specific latent variable $z_S$ is constrained by an information bottleneck. (3) We finetune online with self-generated rewards to master a new domain. (4) We evaluate $\pi$ in $\mathcal{D}_\text{test}$.}
    \label{fig:fig1}
\end{figure}



We study this problem in the setting of learning a policy from vision for performing industrial insertion tasks.
This family of assembly tasks, including plugging in connectors into sockets, keys into locks, screwdrivers into screw intrusions, setting screws, and so on, are found in many stages of manufacturing.
When automated in factories today, these tasks are done by robots with specialized control algorithms that rely on precise localization of the socket location.
For robots to perform these insertion tasks in industrial and warehouse settings with less human supervision, or in unstructured environments such as homes, they must rely on highly accurate state information of the external world (e.g., socket position and in-hand pose estimation).
But such state estimation, using either machine learning or computer vision approaches, is brittle on unseen connectors.
To solve the general problem of inserting a novel connector, one promising solution is to generalize previously collected experience of connector insertion to learn a policy to insert connectors from vision.
Among these tasks, there is enough variability to require generalization and adaptation, but also enough internal structural regularity that we expect transfer between connectors.
We first collected a large offline dataset with insertion data of \numconnectors{} connectors across 2 robots and diverse backgrounds with actions, images, and sparse reward labels.
Offline RL on this data alone generalizes to connectors very similar to those in the training dataset, but we will also expect robots to be able to perform tasks in new domains, perhaps after some practice.
How can a robot insert test connectors from vision in this setting, utilizing offline RL from offline data to enable active online finetuning on a new connector?


The key insight is that we need to (1) adapt to new tasks quickly with online finetuning if the zero-shot solution is not sufficient and (2) generalize to new domains by finding common structure between domains while preserving important domain-specific information.
Ideally, a policy trained offline can generalize from vision to new tasks.
But if it does not, we can still finetune in a new domain with minimal supervision as long as we have a reward function that generalizes instead.
For training policies and reward functions that generalize to test domains, we propose a split representation that combines domain adversarial neural networks~\cite{ganin2016domainadversarial} for domain invariance and a variational information bottleneck~\cite{alemi2017vib}
for controlling the flow of domain-specific information.
This representation, which we call \methodname{} (\methodabbrv) is used first for learning a robust reward function to detect successful insertions for an unseen connector.
Next, we modify implicit Q-learning (IQL), an offline RL algorithm amenable to online finetuning, to use \methodabbrv{}.
During online finetuning, DAIB can be used in combination with online RL to enable fast learning of novel connectors.

We present two main contributions.
We demonstrate a system for finetuning under realistic real-world constraints with minimal human supervision, and applied it to insert connectors robustly from vision without the need of accurate socket localization, both for observations and rewards. 
To accomplish this, we propose a novel representation learning method that allows better generalization of policies and reward functions to unseen domains.
We outperform regression-based baselines on the same dataset that combine localizing the socket with hand-designed control policies, as well as prior RL methods.
We show that new tasks can be finetuned within 200 trials (about 50 minutes of real-world interaction),
given our dataset of off-policy data from \numconnectors{} prior domains of 70,000 trajectories.
This system allows us to finetune IQL to a test connector, increasing performance significantly over the offline performance.
Project videos and our dataset of robotic insertion will be made public at \href{https://sites.google.com/view/learningonthejob}{sites.google.com/view/learningonthejob}


\section{Related Work}




Reinforcement learning has been applied to a variety of robotics tasks~\cite{peters2008baseball, kober2008mp,deisenroth2011pilco, levine2016gps, levine2017grasping, zhu2019hands, giusti15trails, nakanishi2004bipedlfd, kalakrishnan09terraintemplates}.
To utilize offline datasets with diverse data in robotics, algorithms developed for offline RL~\cite{fujimoto19bcq, kumar2020cql, nair2020awac, wu2019brac} have been studied in the robotics setting~\cite{singh2020cog, singh2020parrot, chebotar2021actionable, kalashnikov2021mtopt, kumar2021workflow}.
A subset of offline RL algorithms are amenable to finetuning ~\cite{nair2020awac,villaflor2020finetuning, meng2021starcraft, lee2021finetuning,kostrikov2021iql, liu2022robotlotj}.
Our work builds on the direction of offline pretraining followed by online finetuning in robotics.
But beyond this line of work, we focus on finetuning from visual input in realistic settings with multiple domains and without ground truth reward functions for the new task.

In this respect, our work is closest to prior work on self-supervised RL that does not assume an external reward function and instead learns it from data.
One class of self-supervised RL methods uses goal-conditioned RL with self-supervised rewards~\cite{kaelbling1993goals, schaul2015uva, Baranes2012, andrychowicz2017her, nair2018rig, nachum2018hiro, held2018goalgan, Pere2018, wadefarley2019discern, pong2020skewfit, khazatsky2021val}.
While general, this class of methods is a poor fit for industrial insertion, as high precision is required in both the policy and in evaluating rewards.
Instead, we train a domain generalizing reward classifier from prior data.
Prior methods have used learned rewards~\cite{pong2022smac} and classifier rewards have been proposed as a scalable solution for robotics tasks previously~\cite{fu2018vice, singh2019raq}.
However, learned rewards have not been shown to be useful for finetuning in novel real-world robotic domains previously.
Because we focus on applying offline RL and finetuning from vision in the industrial insertion setting, domain generalization of the reward function is vital for our method to work in practice.

Many aspects of robotic insertion, or peg-in-hole assembly, has been studied in prior work~\cite{whitney1982assembly, xia2006dynamic, zhang2008assembly, zhang2011assembly, li2014usbgelsight, marvel2018insertionsearch}, often utilizing geometry and dynamic analysis, force control, tactile sensing, and search, but these methods can be brittle to state estimation errors.
Learning-based methods, including RL, have also been applied, usually 
for a single connector from ground-truth state information~\cite{luo2019impedance, schoettler2019insertion, lian2021insertionbenchmark}.
In these cases, the RL algorithm must learn to navigate the specific dynamics of the single connector, but does not generalize across connectors.
More recent work has considered using meta-learning to generalize and improve few-shot between domains~\cite{Schoettler2020}.
Zhao et al. use offline RL and finetuning combined with meta-learning to adapt to a new connector~\cite{zhao2022insertion}.
This work assumes a known position of the socket and consistent grasping of the connector, and is robust to a small amount ($\pm1\text{mm}$) of noise.
With known socket position and small error, the learning algorithm can learn a structured noise or exploration strategy that can overcome these errors.
In contrast, we initialize connectors within $\pm20\text{mm}$ of the socket (20$\times$ the variance), which requires the robot to rely on visual feedback since blind exploration will rarely succeed.

Closest to our work is prior work that also uses pixel input for robotic insertion.
Luo et al. incorporate vision alongside proprioception, using a VAE to embed pixel input~\cite{luo2021insertion}.
InsertionNet uses a vision system to localize the object and socket, operating on a "residual policy" which is learned from state by supervised learning~\cite{spector2021insertionnet}.
InsertionNet 2.0 incorporates contrastive representation learning to improve performance~\cite{spector2022insertionnet2}.
These prior works collect data on a single connector and then show robust insertion of that connector.
In contrast, our work focuses on what can be done to leverage prior experience for a novel connector without having access to localization of the socket for supervision.
Our work also demonstrates robustness to larger variation in initial connector pose, up to 20mm error, than prior work.
For visual generalization to a test connector from our offline dataset,
a suitable representation learning method is vital.

Many prior methods have explored representation methods for improving the sample efficiency of RL algorithms, including reconstructive objectives, bisimulation, contrastive methods, latent space prediction, forward model prediction, and other mutual information objectives~\cite{lange2010deep, lange2012autonomous, finn2016deep, ferns2004bisimulation, castro20bisimulation, zhang2021dbc, oord2018cpc, laskin2020curl, nguyen2021tpc, schwarzer2020spr, hafner2018planet, mazoure2020dim, Gregora, Hafner2020, rakelly2021mi, jonschkowski2017pve, ghosh2018learning, sax2018midlevel}.
In this work, the representation learning challenge 
is to be able to generalize to new domains from prior domains for RL finetuning to function.
Thus, most closely related to our work is domain generalization and domain adaptation. Domain adaptation methods generalize from source domains to a target domain, usually by matching the distribution of features between domains via matching statistics or using an adversarial loss~\cite{tzeng2014domainconfusion, sun2016coral, long2015adaptation, ganin2016domainadversarial, Bousmalis2016} and has been applied successfully in the sim-to-real setting~\cite{bousmalis2017simtoreal, james218simtosim}.
Successfully matching distributions makes features indistinguishable between domains.
However, in many settings including in our insertion setting, domain-specific information is also important and full domain invariance is not desired.
We propose a domain generalization method that can trade off between domain-invariant and domain-specific information. 



\section{Problem Statement}


In RL, we consider a MDP $\mathcal{M}$ with states $s_t \in \mathcal{S}$, actions $a_t \in \mathcal{A}$, dynamics $p(s_{t+1}|s_t, a_t)$, and reward $r_t$. The RL agent learns a policy $\pi(a_t|s_t)$ to maximize return $\mathcal{R}_t = \sum_{i=t}^T \gamma^{i-t} r_t$, where horizon $T$ may be infinite, and $\gamma$ is the discount factor.
Many algorithms have been developed for this purpose~\cite{schulman2015trpo, schulman2017ppo, haarnoja2018sac}.
For real-world robot learning, sample efficiency is vital, and algorithms that estimate the value function $V^\pi(s_t) = \E_{p_\pi(\tau)}[\mathcal{R}_t|s_t]$ or action-value $Q^\pi(s_t, a_t) = \E_{p_\pi(\tau)}[\mathcal{R}_t|s_t, a_t] =  \E
_{p(s_{t+1}|s_t, a_t)}
[r(s_t, a_t, r_{t+1}) + V^\pi(s_{t+1})]$ better utilize off-policy data.

In the real world, we often do not want to solve a single static environment but rather learn a policy that can generalize to a variety of new scenarios based on prior experience.
To formalize this, we will consider a set of MDPs $\mathcal{M}_d \sim p(\mathcal{M})$, indexed by the domain index $d$, with shared observation and action spaces but potentially different dynamics and reward~\cite{kirk2021generalizationrl}.
We will consider domains $\{0, 1, \dots, n_d \}$ to be training domains and $d_\text{test}$ be a test domain.
The problem is to maximize returns in a test domain $\mathcal{D}_\text{test}$ with as little as data from the test domain as possible, given experience from the training domains.

From the training domains, we assume access to a prior dataset of $n_d$ domains drawn from a distribution of domains $p(\mathcal{D})$,
with each domain $\mathcal{D}_d = \{\tau_i^d\}_{d=1}^{n_d}$ containing trajectories of transitions $\tau_i^d = \{(s_t, a_t, r_t, s_{t+1})^d\}$.
These trajectories can be of arbitrary quality and used within an offline RL framework to train offline RL algorithms.
The challenge is then to utilize the offline data effectively in order to maximize returns through online interaction in a test domain $\mathcal{D}_\text{test}$.
In the real-world, obtaining a reward on a test domain can be difficult - sometimes as difficult as specifying the optimal policy - as it requires human intervention or supervision.
For this reason, we do not assume we receive a reward signal in $\mathcal{D}_\text{test}$ during online interaction, and reward information must instead be extracted from the training data, where the data collection can be carefully controlled.

\section{Robot Setup and Dataset}
\label{sec:robotsetup}






\begin{wrapfigure}[16]{l}{3.5cm}
    \vspace{1em}
    \centering
        \includegraphics[width=0.99\linewidth]{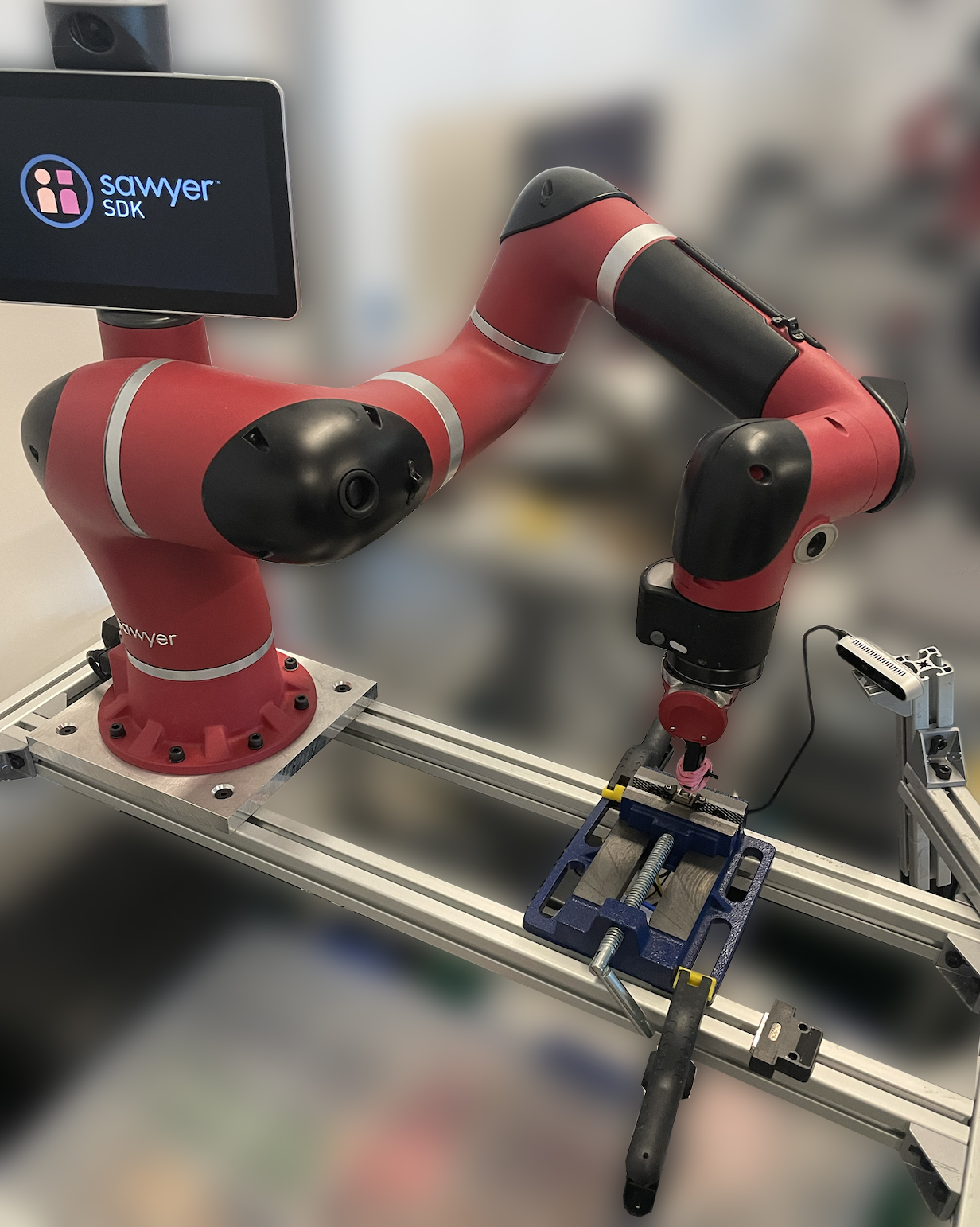}

    \caption{Full view of the robot setup, including a Rethink Sawyer robot, an Intel Realsense D435 camera, and a mount for various connectors.}
    \label{fig:fullrobot}
\end{wrapfigure}
We instantiate real-world learning on the job for the family of industrial insertion tasks, which exhibit real-world challenges that require learning policies from pixels and generalization from few domains.
Individual connectors are pictured in Fig~\ref{fig:fig1} and a third-person view of the robot setup for connector insertion is pictured on the project website.
We use two 7 DoF Sawyer robots for collecting data and running experiments. The robot is commanded with an end-effector controller
in Cartesian coordinates. Desired joint angles are computed via inverse kinematics, and executed with a joint-space velocity controller with force limits for safety and to prevent dislodging connectors from a grasped position.

We initially collect a dataset of insertions of \numconnectors{} connectors, consisting of about 70,000 trajectories total.
Each trajectory is at most 20 steps, but terminates early on a successful insertion.
For each connector, it is grasped by the robot and the socket is mounted to a clamp.
Before data collection, the robot is manually placed in a successful grasp position and a calibration procedure captures the initial pose, which is used for computing exact state position relative to the socket and for detecting ground truth success.
During offline data collection, the robot follows a manual insertion strategy, moving towards the center-line of the socket based on the calibrated pose until within 1mm, then moving downwards.
This expert policy is available during the offline data collection because of manual calibration, which we do not do in the test domain.
Noisy actions are executed with probability $0.2$ to visit a diverse set of states and induce recovery behavior in the prior data.
Sometimes the connector becomes dislodged from the robot grasp and the state-based expert policy fails -- this data is manually discarded.
During this phase, the policy has the benefit of exact relative position between the connector and socket from calibration; the key challenge is that when finetuning to a new connector, this information is not available.


We want to utilize this data to allow the robot to insert a held-out connector not seen during training.
In the online practice phase, a test connector is grasped and mounted to the robot gripper.
Unlike the offline data, the connector position relative to the socket is not made available to the agent, as calibrating it would require human intervention.
This means the agent can actively practice on the new task, but it must be self-supervised: it must estimate its own reward signal and operate from rich visual input.


\section{Learning on the Job with RL Finetuning}

Our method combines offline pretraining and online finetuning to rapidly adapt to new connectors. We will first provide an overview of the RL system, and then describe the representation learning method that we use to enable generalization for both the reward model and policy.

\subsection{System Overview}

We aim to ``learn on the job'' to finetune to $\mathcal{D}_\text{test}$ utilizing data from related training domains $\{\mathcal{D}_1, \mathcal{D}_2, \dots, \mathcal{D}_{n_d}\}$. We can interact in $\mathcal{D}_\text{test}$ but without rewards, so we need to extract artifacts from the training data such that we can coherently practice and master the task in the test domain.


Our solution is illustrated in Fig.~\ref{fig:fig1} and can be considered in four phases.
(1) We collect an offline dataset on a diverse set of connectors.
Our data collection procedure for the insertion setting is described in Section~\ref{sec:robotsetup} above.
(2) In the offline training phase, we run offline RL to train a policy, value function, and reward model $R_\psi(s)$ on the offline dataset.
These models all take images as input, and the training details are discussed in Section~\ref{sec:methodofflinerl}.
We discuss techniques to improve generalization to new domains in Section~\ref{sec:reward}. 
(3) In the online phase conducted interactively in the test domain with minimal human supervision, we start with the pre-trained policy and value function, and finetune the policy and value function with online data.
Since we don't have rewards in the test domain, we use $R_\psi(s)$ to obtain reward predictions.
(4) In the evaluation phase, we measure the ability of the finetuned policy to succeed at the task. Success rates are measured manually, as we do not assume ground truth rewards in the test environment.

\subsection{Offline Training Phase}
\label{sec:methodofflinerl}

In the offline training phase, we wish to recover artifacts that can be used to maximize performance on the test domain.
One option is offline RL: a variety of methods have been developed for the purpose of training policies and value functions from offline data.
However, offline policies may not consistently generalize to a test domain.
Since we can interact actively with the test domain, the alternative is to treat the offline phase as preparation for online finetuning.

A subset of offline RL algorithms are amenable to online finetuning.
In this work, we use IQL~\cite{kostrikov2021iql}, which
learns value functions by quantile regression, optimizing:
\begin{eqnarray}
    \mathcal{L}_{Q} &=&  \mathbb{E}_{(s, a, s') \sim \mathcal{D}} \left[(r(s) + \gamma \bar{V}(s') - Q(s, a))^2 \right] \label{eqn:loss_q}  \\
    \mathcal{L}_{V} &=& \mathbb{E}_{(s, a) \sim \mathcal{D}} \left[L_2^\tau(\bar{Q}(s, a) - V(s))^2 \right], \label{eqn:loss_v}
\end{eqnarray}
where $L_2^\tau(u) = |\tau-\mathds{1}(u < 0)|u^2$, and $\overline{\cdot}$ indicates a stop gradient. The policy is extracted from the value function with advantage-weighted regression~\cite{peng2019awr}:
\begin{eqnarray}
    \mathcal{L}_{\pi} &=& 
    \enspace 
    \mathbb{E}_{(s, a) \sim \mathcal{D}} \left[\log \pi(a|s) \exp(\bar{A}(s, a)/\beta) \right], \label{eqn:loss_pi}
\end{eqnarray}
where $A(s, a) = Q(s, a) - V(s)$ and $\beta$ is a temperature parameter. Each function, $\pi$, $Q$, and $V$ is parametrized as a CNN that take images as state input, using independent ResNet-18 backbones~\cite{he2016resnet}.

The trained policy could in theory be applied to any test connector. However, when faced with a new connector, the offline-trained policy may not generalize well in zero shot.
In that case, we want to be able to actively finetune the policy to solve the task through additional trials with the new connector.
To do so, we will also require a reward function computed from raw image observations, as the robot does not have access to the socket position (and therefore rewards) in the test domain.
In the insertion dataset, the offline data contains binary rewards, so we train a reward classifier $R_\psi(s)$ with binary cross entropy:
$\mathcal{L}_\text{rew}(R; s, r) = -r\log(R_\psi(s)) - (1-r)\log(1-R_\psi(s))$

\subsection{Self-Supervised Online Finetuning Phase}

Next, in the online finetuning phase, we collect trajectories using the pretrained policy with additional exploration noise.
In this phase, we do not assume a ground truth reward, as this requires accurate state estimation and localization.
Instead, we evaluate the reward from the reward model.
For a transition $(s, a, s')$, we compute the predicted reward $\hat{r} = R_\psi(s')$ and append $(s, a, s', \hat{r})$ to the replay buffer.
We then run the batch gradient descent updates of the policy, Q function, and value function according to equations~\ref{eqn:loss_q},~\ref{eqn:loss_v}, and~\ref{eqn:loss_pi}.
To balance the online and offline training data, we separate the data into unique replay buffers and in each mini-batch we sample $25\%$ online data, with no data augmentations, and $75\%$ offline data with data augmentations.


\subsection{Domain Adversarial Information Bottleneck}
\label{sec:reward}

This system for offline RL and online finetuning can in theory finetune to novel domains, but the degree to which it succeeds depends heavily on the generalization of the offline policy, value functions, and rewards to the test domain.
As we show in our ablation experiments, this base system often fails to solve test tasks within a practical time frame on the robot.
Therefore, we need a method that generalizes more effectively.
In this section, we will describe how to use the domain adversarial information bottleneck to learn an intermediate representation $z=g_\phi(s)$ to improve generalization of the reward model, then explain how we incorporate this bottleneck into RL in the following section.
We will overload notation in this section, referring to $R_\psi(g_\phi(s))$ as $R_\psi(z)$.

The representation should allow us to generalize to an unseen domain by capturing the common structure from previously seen domains.
So what common structure does our insertion domain have?
Coarsely, these tasks are peg insertion variants.
At a finer level, the structural similarity of the reward function and expert policy to solve these tasks is clustered into groups of connectors ---{} for instance, NEMA connectors or Deutsch DT series as shown in Figure~\ref{fig:fig1}.
Within each group, the visual features for aligning the connector with the socket and the insertion depth is shared.

The most commonly proposed approach for generalizing to new domains is domain invariance~\cite{ganin2016domainadversarial}.
In our setting, a fully domain invariant representation could not take advantage of domain-specific information about the similarity between certain connectors: at which depth the reward is obtained, connector-specific visual cues that the policy or reward classifier can take advantage of, and so on.
Instead, consider a split $z = (z_I, z_S)$ consisting of a domain invariant representation $z_I$ and a domain specific representation $z_S$. Can we design an objective to factorize the two representations, such that it improves generalization to new domains?

For enforcing domain invariance of $z_I$, we take inspiration from domain adversarial neural networks (DANN)~\cite{ganin2016domainadversarial}, which backpropagates the signal from a domain classifier into the representation. The domain classifier $F_\theta$ is trained to minimize negative log likelihood:
$\mathcal{L}_\text{domain}(F; s, d) = -\log F_\theta(z_I)_d.$
For training the reward model, $\mathcal{L}_{adv} = -\mathcal{L}_\text{domain}(F; s, d)$ is added as an auxiliary objective, adversarially optimizing $z_I$ to worsen the domain classifier.

The auxiliary loss imposes a cost for any domain specific information in $z_I$, but as discussed earlier, allowing for some domain-specific information may be important for the classifier to perform well.
However, if we simply concatenate a representation $z_S$ without a domain invariance loss, $z_S$ carries all the information and the features $z_I$ degenerate to be completely uninformative when trained.
As we show in our experiments, this 
leads to reduced performance, similar to ERM in theory and in practice since $z_I$ is not used. 
We need to somehow limit the information carried by $z_S$, such that the model maximizes the domain-invariant processing captured by $z_I$. A natural tool to accomplish this is the variational information bottleneck (VIB)~\cite{alemi2017vib}.
To learn a split representation, we constrain the information through $z_S$:
\begin{eqnarray}
\mathcal{L} = \mathcal{L}_{\text{rew}}(z) + \mathcal{L}_{adv} \; \text{s.t.} \; I(s; z_S) \leq C.
\end{eqnarray}

Following Alemi et al.~\cite{alemi2017vib}, we can turn this into an unconstrained problem and compute the evidence lower bound with encoder $q_\phi(z_S|s) = \mathcal{N}(\mu_\phi(s), \sigma_\phi^2(s))$:
\begin{eqnarray}
\mathcal{L_\text{R}} = \E_{\epsilon \sim p(\epsilon)} \mathcal{L}_\text{rew}(z) + \underbrace{\mathcal{L}_{adv} + \lambda[KL(q_\phi(z_S|s)||p(z_S))]}_{\mathcal{L}_\text{DAIB}(z)}
\end{eqnarray}

The prior $p(z_S)$ is taken to be an isotropic Gaussian. We use $\lambda=0.01$ and optimize this objective to learn a reward classifier and representation with Adam~\cite{kingma2014adam}.
The architecture and losses are pictured in Fig.~\ref{fig:fig1} in the offline training section.
Further details are provided on the website.


\subsection{Reinforcement Learning with DAIB}

Besides the reward generalizing to a new domain, we also need the policy to generalize at least partially to a new domain to obtain meaningful exploration data to improve from with finetuning.
We incorporate a domain adversarial information bottleneck into the policy to enable domain generalization.
Let $z^\pi = g_\phi^\pi(s)$ represent the output of the CNN backbone of the policy.
We optimize the following policy loss:
    $\mathcal{L}_{\pi} + \mathcal{L}_\text{DAIB}(z^\pi).$
We find that using DAIB for the policy bottleneck enables consistent performance across connectors, as explained further in Section~\ref{sec:finetuningexps}.

In the online finetuning phase, when we sample an action from the policy during a rollout and recompute the reward from the reward model, we use a deterministic encoding in the information bottleneck, where the mean of the output distribution is taken as the sampled value.
We freeze the convolutional layers during the online finetuning phase.
We refer to the overall system
as ``learning on the job'' (LOTJ).

\section{Experiments}

\renewcommand{\arraystretch}{1.4}
\setlength{\arrayrulewidth}{0.1mm}
\setlength{\tabcolsep}{8pt}

\begin{table*}[]
\begin{tabular}{l|r|r|r|r|rr|rr|rr}
Connector & \multicolumn{1}{l}{Localize} & \multicolumn{1}{l}{\begin{tabular}[c]{@{}l@{}}Straight\\ Down\end{tabular}} & \multicolumn{1}{l}{\begin{tabular}[c]{@{}l@{}}Random\\ Search\end{tabular}} & \multicolumn{1}{l}{\begin{tabular}[c]{@{}l@{}}Spiral\\ Search\end{tabular}} & \multicolumn{1}{l}{\begin{tabular}[c]{@{}l@{}}SAC \\ (offline)\end{tabular}} & \multicolumn{1}{l}{\begin{tabular}[c]{@{}l@{}} \\ (online)\end{tabular}} & \multicolumn{1}{l}{\begin{tabular}[c]{@{}l@{}}State-IQL \\ (offline)\end{tabular}} & \multicolumn{1}{l}{\begin{tabular}[c]{@{}l@{}} \\ (online)\end{tabular}} & \multicolumn{1}{l}{\begin{tabular}[c]{@{}l@{}}LOTJ, full\\ (offline)\end{tabular}} & \multicolumn{1}{l}{\begin{tabular}[c]{@{}l@{}}\\ (online)\end{tabular}} \\ \hline \hline
DT12-1    & 3/20                         & 1/20                                                                        & 7/20                                                                        & 4/20                                                                          & 0/20                                                                          & 0/20                                                                          & 5/20                                                                          & 5/20                                                                        & 7/20                                                                         & \textbf{19/20}                                                                 \\
Mega 1x1  & 3/20                         & 1/20                                                                        & 1/20                                                                        & \textbf{15/20}                                                                          & 0/20                                                                          & 0/20                                                                          & 0/20                                                                          & 2/20                                                                       & 5/20                                                      & \textbf{16/20}                                                  \\
NEMA 15   & 0/20                         & 1/20                                                                        & 0/20                                                                        & 3/20                                                                            & 0/20                                                                          & 0/20                                                                          & 4/20                                                                          & 1/20                                                                      & \textbf{11/20}                                                                        & \textbf{15/20}                                                                 \\
Europlug  & 6/20                         & 4/20                                                                        & 6/20                                                                        & 4/20                                                                            & 0/20                                                                          & 0/20                                                                          & 6/20                                                                          & 4/20                                                                      & \textbf{10/20}                                                                        & \textbf{15/20}                                                                
\end{tabular}
\caption[LOF]{\footnotesize Insertion scenario I: the initial position of the gripper is randomly offset by $\pm$10mm from above the socket. The best performing method in each row is bolded. (Statistical significance is computed according to Barnard's exact test~\cite{mehrotra2003testing} for independent binomials at $95\%$ confidence level.) Only our method online after fine-tuning solves all four test tasks the majority of the time. }
\vspace{-0.3cm}
\label{table:easy}
\end{table*}


\renewcommand{\arraystretch}{1.4}
\setlength{\arrayrulewidth}{0.1mm}
\setlength{\tabcolsep}{8pt}

\begin{table*}[]
\begin{tabular}{l|r|r|r|r|rr|rr|rr}
Connector     & \multicolumn{1}{l}{Localize} & \multicolumn{1}{l}{\begin{tabular}[c]{@{}l@{}}Straight\\ Down\end{tabular}} & \multicolumn{1}{l}{\begin{tabular}[c]{@{}l@{}}Random\\ Search\end{tabular}} & \multicolumn{1}{l}{\begin{tabular}[c]{@{}l@{}}Spiral\\ Search\end{tabular}} & \multicolumn{1}{l}{\begin{tabular}[c]{@{}l@{}}SAC \\ (offline)\end{tabular}} & \multicolumn{1}{l}{\begin{tabular}[c]{@{}l@{}} \\ (online)\end{tabular}} & \multicolumn{1}{l}{\begin{tabular}[c]{@{}l@{}}State-IQL \\ (offline)\end{tabular}} & \multicolumn{1}{l}{\begin{tabular}[c]{@{}l@{}} \\ (online)\end{tabular}} &
\multicolumn{1}{l}{\begin{tabular}[c]{@{}l@{}}LOTJ, full\\ (offline)\end{tabular}} & \multicolumn{1}{l}{\begin{tabular}[c]{@{}l@{}}\\ (online)\end{tabular}} \\ \hline \hline
DT12-1        & 0/20                         & 0/20                                                                        & 0/20                                                                        & 0/20                                                                        & 0/20                                                                             & 0/20                                                                          & 0/20                                                                          & 3/20                                                                          & 5/20                                                                      & \textbf{18/20}                                                                 \\
Mega 1x1 & 0/20                         & 1/20                                                                        & 0/20                                                                        & \textbf{17/20}                                                                             & 0/20                                                                          & 0/20                                                                          & 0/20                                                                          & 11/20                                                                    & 0/20                                                                         & \textbf{20/20}                                                                 \\
NEMA 15     & 0/20                         & 0/20                                                                        & 0/20                                                                        & 0/20                                                                             & 0/20                                                                          & 0/20                                                                          & 3/20                                                                          & 1/20                                                                     & 6/20                                                                         & \textbf{17/20}                                                                 \\
Europlug      & 0/20                         & 0/20                                                                        & 0/20                                                                        & 0/20                                                                            & 0/20                                                                          & 0/20                                                                          & 0/20                                                                          & 5/20                                                                      & 2/20                                                                         & \textbf{15/20}                                                                
\end{tabular}
\caption[LOF2]{\footnotesize Insertion scenario II: the initial position of the gripper is sampled from a box 10-20mm from the socket. This scenario is more difficult as moving straight down almost never solves the task. The baselines in this series of experiments use the localization model initially, then follow the baseline strategy. Our method significantly improves from offline performance on all connectors. }
\label{table:hard}
\vspace{-0.5cm}
\end{table*}

\renewcommand{\arraystretch}{1.4}
\setlength{\arrayrulewidth}{0.1mm}
\setlength{\tabcolsep}{3pt}

\renewcommand{\arraystretch}{1.4}
\setlength{\arrayrulewidth}{0.1mm}
\setlength{\tabcolsep}{3pt}

\begin{table}[]
\centering
\begin{tabular}{l|c|c|c}
         & LOTJ, no DAIB & LOTJ, DAIB reward & LOTJ, full                                \\ \hline \hline
DT12-1   & 0/20 $\rightarrow$ 5/20           & 4/20 $\rightarrow$ 0/20               & 0/20 $\rightarrow$ 18/20 \\
Mega 1x1 & 3/20 $\rightarrow$ 0/20           & 3/20 $\rightarrow$ 20/20              & 0/20 $\rightarrow$ 20/20                     \\
NEMA 15  & 0/20 $\rightarrow$ 0/20           & 0/20 $\rightarrow$ 2/20               & 6/20 $\rightarrow$ 17/20                     \\
Europlug & 0/20 $\rightarrow$ 15/20          & 1/20 $\rightarrow$ 12/20              & 2/20 $\rightarrow$ 15/20                    
\end{tabular}
\caption{\footnotesize Ablation of using the domain adversarial information bottleneck (DAIB) for the reward and for the policy across all four test connectors. The only consistent setting where finetuning occurs across all four connectors is using the DAIB for both reward and policy. Removing the bottleneck for the policy significantly reduces finetuning performance on two tasks, but still finetunes on the other two. Removing the bottleneck for the reward prevents finetuning except for the Europlug connector. }
\vspace{-0.5cm}
\label{table:daibablation}
\end{table}

In our experiments, we first evaluate the effectiveness of the proposed system at inserting novel connectors, and compare it with a variety of baselines. Then, we ablate the method and DAIB in particular to isolate the contribution of the representation learning component.

\subsection{Finetuning Comparisons}
\label{sec:finetuningexps}


For insertion of a novel connector, we first run offline training to obtain a policy, Q function, and value function.
Then, we run online finetuning of RL on the novel connector.
We evaluate four connector insertion tasks: Deutsch DT 12-way, Megablock, NEMA 15-5, and Europlug.
In practice, we would simply train offline once and apply it to any new connector, but for statistically rigorous results on several test connectors, we train offline on a set of connectors excluding that connector or close variants (e.g., same connector on a different robot), similar to hold-one-out cross-validation.
We evaluate in two settings: an easier scenario where the initial location of the connector is centered around the socket with $\pm10\text{mm}$ noise, and a harder scenario where the initial location of the connector is offset by $10-20\text{mm}$.

In the easier $\pm10\text{mm}$ setting, we compare against four hand-engineered baselines and two RL baselines. 
\textbf{Localize}: represents a method with learned perception and hand-designed control, where we train a model (on the same offline data) to predict the socket position. The controller regresses onto the state positions in the same offline data at each step and executes a hand-engineered
policy to reach that goal location.
\textbf{Straight down}: moves straight down from the starting position.
\textbf{Random search}: executes a stochastic search policy as described in~\cite{marvel2018insertionsearch}, moving straight down from the initial starting position and then move to 5 randomly sampled positions while pressing down.
\textbf{Spiral search}: moves in a spiral while pressing downward, as described in~\cite{marvel2018insertionsearch}.
\textbf{SAC}: uses soft actor critic from vision~\cite{haarnoja2018sac} as the underlying RL algorithm, similar to~\cite{schoettler2019insertion}, and does not use DAIB.
\textbf{State-IQL}: trains IQL from state instead of images during offline training, then uses the localization model for state estimation during online training.
Each method including ours is executed for 20 time steps per trajectory for evaluation.
For the RL methods, we report offline performance after offline training and then online performance after finetuning.
The policies were finetuned for 200 trials, or about 50 minutes.


The results are reported in Table~\ref{table:easy}. We see that the performance of most methods is inconsistent, but LOTJ online
successfully solves the task to $>75\%$ success for all four new connectors.
In the harder 10-20mm setting, we compare against similar baselines. However, we found that the simpler straight down, random search, and spiral search methods always fail from the initial position, as it is too far from the socket. Instead, we first execute the localization model and move to that goal position, then run the corresponding strategy.
The results for the harder insertion scenario are reported in Table~\ref{table:hard}. In this case, the performance of most methods, including LOTJ offline,
on most connectors, is poor.
However, LOTJ is able to solve these tasks after 200 trials of online finetuning.
Importantly, even when the initial performance is 0/20, as in the DT 12-way connector where the initial policy is poor and does not get close to the socket, the finetuned process quickly improves the policy and makes contact with the socket within a few trials.
Further trials are required for finetuning to actually observe successes through stochastic exploration to perfect the policy.







\subsection{Domain Adversarial Information Bottleneck Ablations}

\textbf{Reward classification.} To learn a generalizable reward function and to evaluate the proposed domain generalization method, we first need to learn a reward classifier from the offline data. 
The accuracy of this reward classifier has a significant impact on whether online finetuning with the learned reward successfully solves a task; if the reward is inaccurate, the policy can adversarially learn to visit regions of incorrect high reward.
We train the reward classifier on a subset of 25 connectors and evaluated on a set of five held-out connectors.
During training and evaluation, the data is rebalanced to be $50\%$ positive and $50\%$ negative samples per connector.
In Table~\ref{tab:reward}, we report the mean accuracy of each method averaged across the five held-out connectors.

We compare the following methods.
Our method, \textbf{DAIB}, uses a domain adversarial objective in combination with a variational information bottleneck as detailed in section~\ref{sec:reward}.
\textbf{DANN} enforces domain invariance using a domain adversarial neural network~\cite{ganin2016domainadversarial}. \textbf{DAIB, $\lambda=0$} concatenates $z_S$ to the representation as in DAIB, but without enforcing an information bottleneck.
\textbf{VIB} enforces a variational information bottleneck only without domain invariance~\cite{alemi2017vib}.
\textbf{ERM} has no auxiliary representation learning objectives.

\begin{wraptable}[12]{r}{3.0cm}
\begin{tabular}{l|l}
Method   & Acc. \\ \hline
DAIB (Ours) & \textbf{88\%}     \\
DANN          & 71\%     \\
DAIB, $\lambda=0$    & 77\%     \\
VIB         & 79\%     \\
ERM             & 76\%    
\end{tabular}
\caption{Comparison of test accuracy on reward classification.}
\label{tab:reward}
\end{wraptable}
The results are presented in Table~\ref{tab:reward}. We see that our method, DAIB, is able to achieve an $88\%$ accuracy, significantly higher than other methods.
DANN, which enforces domain invariance, is the worst performing. This poor performance shows that the domain invariance assumption is too strong for the task of reward classification, and connector-specific information is likely necessary to identify a successful insertion.
The DAIB, $\lambda=0$
method achieves similar accuracy as ERM, since the network can just make the domain invariant features degenerate and bypass the domain adversarial objective.
The VIB alone slightly improves performance over ERM because of a regularizing effect. But all methods achieve significantly worse accuracy than DAIB. Next, armed with an accurate reward model, we investigate finetuning connectors using the learned reward model.

\textbf{RL ablation.} Next, we ablate using LOTJ for the reward and policy. The results are presented in Table~\ref{table:daibablation} on the harder insertion scenario (10-20mm). The results vary somewhat connector to connector, but using DAIB for both the reward and policy consistently finetunes successfully while removing the bottleneck sometimes fails to finetune.
Visualizations of saliency maps~\cite{selveraju2022gradcam} of trained networks to better understand what features the networks are paying attention to are available on the project website.

\begin{figure}
    \centering
    \begin{subfigure}[b]{0.99\linewidth}
        \center
        \includegraphics[width=0.99\textwidth]{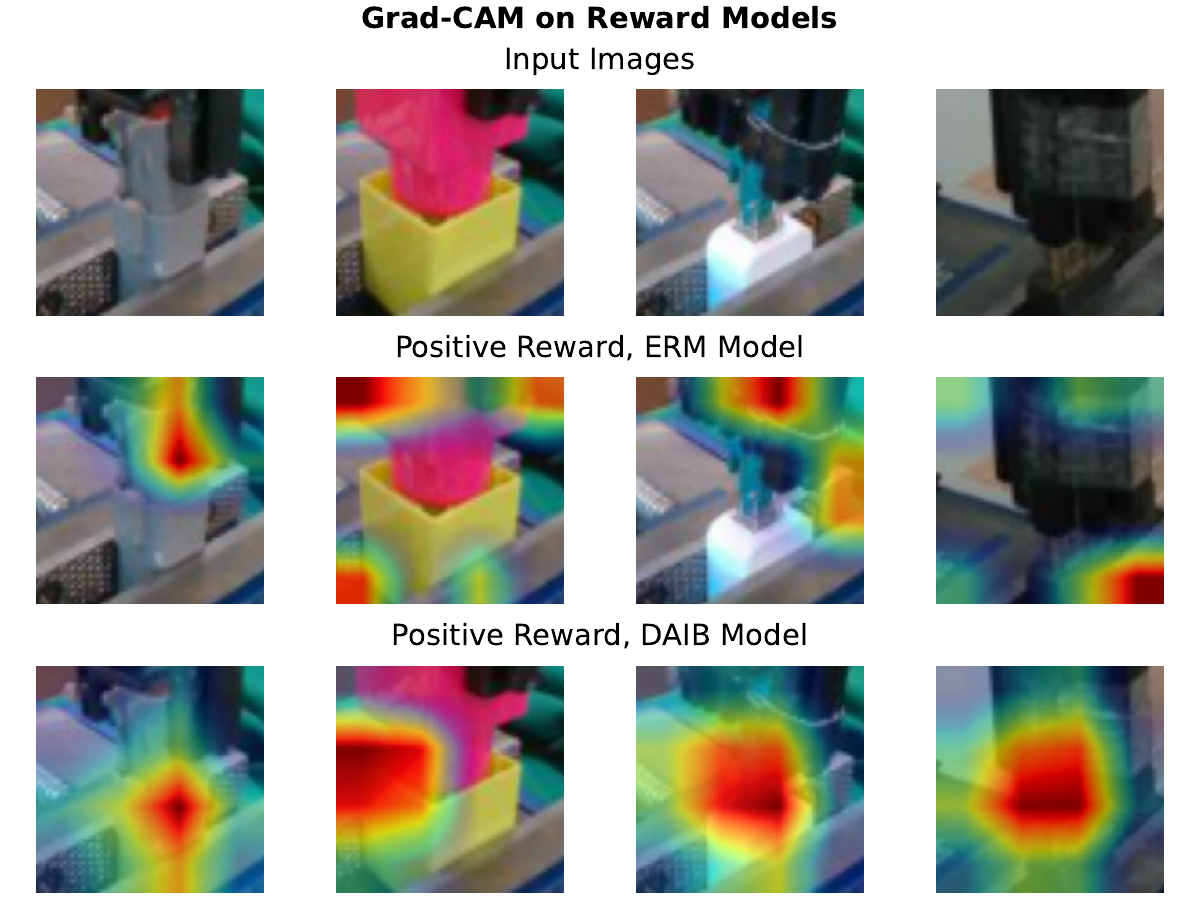}
    \end{subfigure}

    \caption{Grad-CAM visualization of reward models. The first row shows test input images, which all have ground truth reward 1. The next two rows show the Grad-CAM heat map of a trained reward classifier trained with standard ERM and with DAIB. The classifier trained with DAIB focuses on semantically meaningful regions of the connector and socket, while the classifier trained without DAIB often pays attention to spurious regions of the image. }
    \label{fig:gradcam-reward}
    \vspace{-0.5cm}
\end{figure}

\subsection{Inspecting Vision Networks}

In this section, we inspect the learned networks to better understand how the bottleneck and finetuning function.
To do so, we use Grad-CAM~\cite{selveraju2022gradcam}, which was developed for visualizing regions of the image that a convolutional network pays attention to and has been shown to provide sane saliency explanations~\cite{adebayo2018saliency, michaud2020reward}.
For a particular network scalar output $y$ and intermediate feature maps $A^k$, Grad-CAM computes weights $\alpha^k = \frac{1}{Z} \sum_{ij} \partial y / \partial A$ and then computes the 2D saliency mask $\text{ReLU}(\sum_k \alpha^k A^k)$.
A location with higher value in the mask can be interpreted as having more “importance” as it corresponds to a larger gradient, indicating that the location contributes more to a higher value of $y$, often a logit for a classification problem.
We take block 3, layer 1 of the ResNet model as $A$, which has spatial resolution $6\times 6$, and overlay the mask as a heat-map over the input image with bilinear interpolation.

\begin{figure}
    \centering
    \begin{subfigure}[b]{0.99\linewidth}
        \center
        \includegraphics[width=0.99\textwidth]{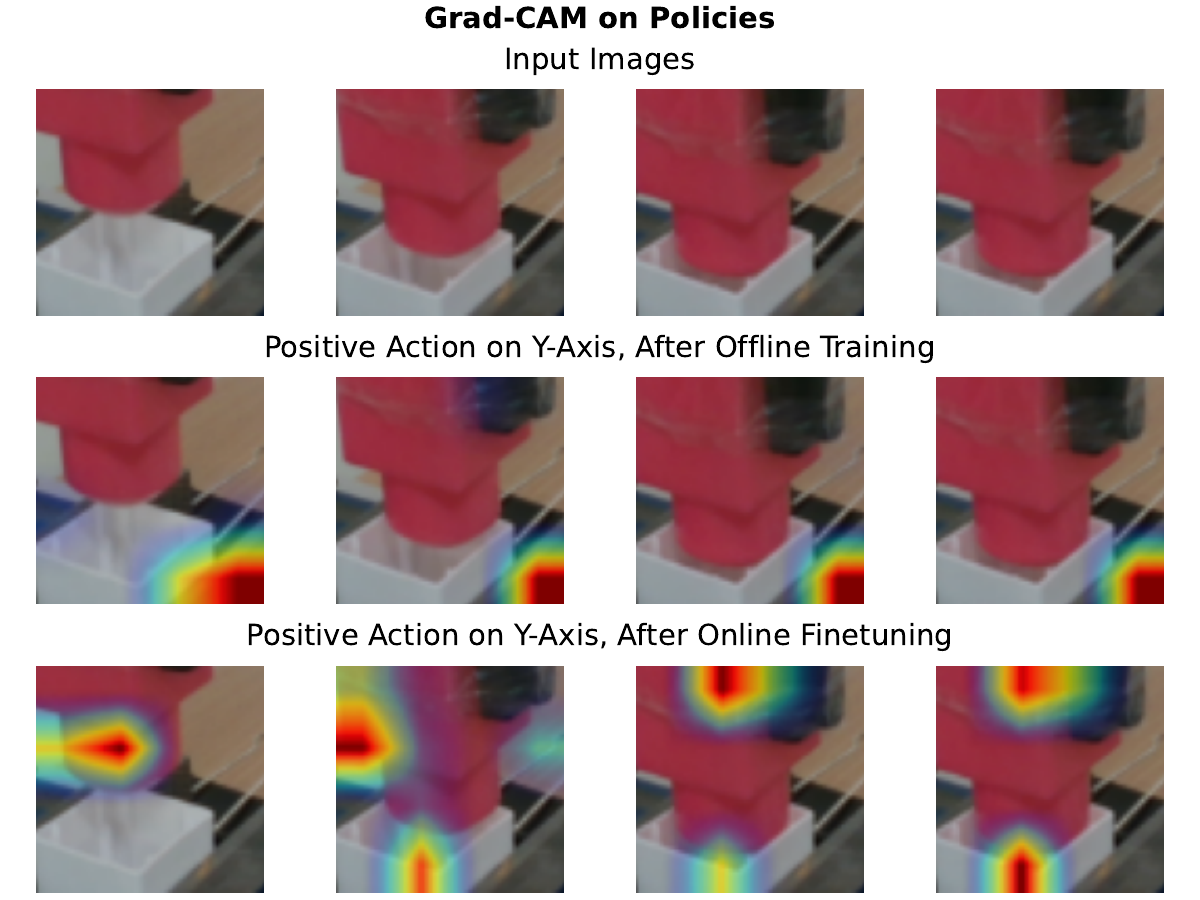}
    \end{subfigure}

    \caption{Grad-CAM visualization of the policy before and after fine-tuning. The first row shows input images from a single trajectory. The next row shows a Grad-CAM heatmap of the Y-axis policy output after offline training, before being trained on any examples of the test connector. The policy is paying attention to a spurious corner of the image. The final row shows the Grad-CAM heatmap after fine-tuning; the policy now attends to the connector and socket positions.
    }
    \label{fig:gradcam-policy}
    \vspace{-0.5cm}
\end{figure}

We first study the effect of the DAIB bottleneck, by visualizing Grad-CAM heat maps on the reward classifier. We visualize which regions of the image the reward classification model pays attention to, to output a positive reward classification. 
We compare a model trained with DAIB and without any additional representation objectives (ERM). 
Figure~\ref{fig:gradcam-reward} shows the results. 
We see from the heat-map overlays that the DAIB model is most influenced by regions with the connector and socket, while the ERM model is influenced most by surrounding regions in the environment such as the gripper or the bench.
The regularizing effect of the DAIB objectives is key to reducing the tendency of the reward model to overfit to details of the environment when training with relatively few domains.
A more complete set of visualizations, including visualizing the policy before and after finetuning, are available on the \href{https://sites.google.com/view/learningonthejob}{website}.

Next, we study the effect of finetuning on the policy. The policy is not a classifier; instead, we compute the gradient of the action in the Y-axis, which is parallel to the edges of the bench grips (back-right to front-left in the view of the camera). The visualized heatmaps effectively tell us: which parts of the image influenced the policy to move to the left?
These visualizations are pictured in Figure~\ref{fig:gradcam-policy} for frames from a single trajectory.
The offline policy (second row) is originally fixated on the bottom right corner of the image, which is uninformative.
After finetuning (third row), the policy instead focuses on the connector and socket.
This highlights the flexibility of the policy in unknown environments given an accurate reward signal, alleviating the burden of zero-shot generalization.

\section{Discussion}

We proposed a system for real-world finetuning of RL policies for connector insertion, which can enable a robot to learn to insert a new connector with minimal human supervision directly from raw image observations. Our system utilizes learned reward classifiers and offline RL for pretraining of generalizable policies. Generalization of both the reward functions and policies is facilitated by a novel domain generalization method, the domain adversarial information bottleneck (DAIB).
This scheme can generally be incorporated in many scenarios where robot datasets contain highly correlated data from related domains -- a common scenario due to hardware and environment setup costs.

One limitation of the method is its strong reliance on the reward model. Our approach is predicated on the hypothesis that it is easier to obtain a generalizable reward model than a generalizable policy -- i.e., it is easier for a robot to figure out if it is succeeding on a new connector than it is for it to actually perform the task. Our experiments show that this generally works well in our experiments. However, if the reward model generalizes incorrectly, for example by producing false positives, that will throw off the subsequent policy finetuning. More careful application of domain adaptation methods that incorporate statistics of the online finetuning process might alleviate this issue.
The reward model can also be improved in an active manner by incorporating human feedback for selected trajectories~\cite{christiano2017feedback, lee2021pebble}.

We collected the offline dataset with a hand-engineered policy in this work; either manually designed policies~\cite{dasari2019robonet} or human demonstrations~\cite{ebert2021bridge} have been previously used for collecting large robotic datasets.
However, for truly large-scale diverse data collection, manually designing policies or requiring a person to provide demonstrations can be very cumbersome.
One promising direction that ``learning on the job'' enables is low-supervision collection of high-quality diverse data, which can be utilized within a lifelong learning framework for either insertion or for general robotics.


\section{Acknowledgements}

We thank Ilya Kostrikov, Kuan Fang, and Patrick Yin for useful discussions. This research was partially supported by Siemens and the Office of Naval Research.

{ \small
\bibliographystyle{IEEEtran}
\bibliography{example}
}

\end{document}